\documentclass[11pt,twoside]{article}
\usepackage{geometry}
\usepackage{fancyhdr}
\fancypagestyle{firststyle}
{
   \fancyhf{}
   \fancyfoot{}
}

\usepackage{amsmath,amssymb,natbib,graphicx,url}
\usepackage[ruled]{algorithm2e}

\usepackage{longtable}

\usepackage{booktabs}

\usepackage{subcaption}

\newcommand{\Order}{\mathcal{O}}
\DeclareMathOperator{\sign}{sign}

\usepackage[utf8]{inputenc}
\usepackage[english]{babel}

\usepackage{rotating}
\usepackage{paralist}

\newtheorem{theorem}{Theorem}

\begin{document}

\title{Multi-Merge Budget Maintenance\\for Stochastic Gradient Descent SVM Training}
\author{Sahar Qaadan and Tobias Glasmachers\\Institut f{\"u}r Neuroinformatik, Ruhr-Universit\"at Bochum, Germany\\\texttt{sahar.qaadan@ini.rub.de, tobias.glasmachers@ini.rub.de}}
\date{}

\maketitle

\begin{abstract}
Budgeted Stochastic Gradient Descent (BSGD) is a state-of-the-art
technique for training large-scale kernelized support vector machines.
The budget constraint is maintained incrementally by merging two points
whenever the pre-defined budget is exceeded. The process of finding
suitable merge partners is costly; it can account for up to $45\%$ of
the total training time. In this paper we investigate computationally
more efficient schemes that merge more than two points at once. We
obtain significant speed-ups without sacrificing accuracy.
\end{abstract}

\section{Introduction}

The Support Vector Machine (SVM; \cite{cortes1995support}) is a 
widespread standard machine learning method, in particular for binary
classification problems.
Being a kernel method, it employs a linear algorithm in an implicitly
defined kernel-induced feature space \cite{Vapnik:95}. 
SVMs yield high predictive accuracy in many applications \cite{Noble:2004,Quinlan:2004,Lewis:2006a,Cui:2007,Son:2010,Yu:2010,Lin:2014,Yu:2016}. 
They are supported by strong learning theoretical guarantees \cite{Joachims:98,Caruana:2006,Glasmachers:2006,Bottou:2007,Mohri:2012,Cotter:2013,Hare:2016}. 
When facing large-scale learning, the applicability of support vector
machines (and many other learning machines) is limited by its
computational demands. Given $n$ training points, training an SVM with
standard dual solvers takes quadratic to cubic time in $n$
\cite{Bottou:2007}. \cite{steinwart2003sparseness} established that the 
number of support vectors is linear in $n$, and so is the storage
complexity of the model as well as the time complexity of each of its
predictions. This quickly becomes prohibitive for large $n$, e.g., when
learning from millions of data points.
Due to the prominence of the problem, large number of solutions were
developed. Parallelization can help \cite{Zanni:2006,Zhu:2009}, but it 
does not reduce the complexity of the training problem.
One promising route is to solve the SVM problem only locally, usually
involving some type of clustering \cite{zhang2006svm,ladicky2011locally} or with a hierarchical divide-and-conquer strategy \cite{graf:2005,hsieh2014divide}. An alternative approach is to leverage the progress in the domain of
linear SVM solvers \cite{Joachims:99,Zhang:2004,Hsieh:2008,Teo:2010}, which scale well to large data sets. To this end, kernel-induced feature
representations are approximated by low-rank approaches
\cite{fine2001efficient,Rahimi:2007,zhang2012scaling,Yadong:2014,Lu:2016}, either a-priory using random
Fourier features, or in a data-dependent way using Nystr{\"o}m sampling.
Budget methods, introducing an a-priori limit $B \ll n$ on the number of support vectors \cite{Nguyen:2005,Dekel:2006,Wang:2012}, go one step further by letting the optimizer adapt the feature space approximation during operation to its needs. This promises a comparatively low
approximation error. The usual strategy is to merge support vectors at
need, which effectively enables the solver to move support vectors
around in input space. Merging decisions greedily minimize the
approximation error.

In this paper we propose a simple yet effective computational
improvement of this scheme. Finding good merge partners, i.e., support
vectors that induce a low approximation error when merged, is a rather
costly operation. Usually, $\Order(B)$ candidate pairs of vectors are
considered, and for each pair an optimization problem is solved with an
iterative strategy. Most of the information on these pairs of vectors is
discarded, and only the best merge is executed. We propose to make
better use of this information by merging more than two points. The main
effect of this technique is that merging is required less frequently,
while the computational effort per merge operation remains comparable.
Merging three points improves training times by $30\%$ to $50\%$, and
merging 10 points at a time can speed up training by a factor of five.
As long as the number of points to merge is not excessive, the same
level of prediction accuracy is achieved.

\section{Support Vector Machine Training on Budget}

In this section we introduce the necessary background: SVMs for binary
classification, and training with stochastic gradient descent (SGD) on a
budget, i.e., with a-priori limited number of support vectors.

\subsection{Support Vector Machines}

An SVM classifier separates two classes by means of the large margin
principle applied in a feature space, implicitly induced by a kernel
function  $k : X \times X \to R$ over the input space 
$X$. For labels
$Y = \{-1, +1\}$, the prediction on $x \in X$ is computed as
\begin{align*}
	\sign \Big( \big\langle w, \phi(x) \big\rangle + b \Big) = \sign \left( \sum_{j=1}^n \alpha_j k(x_j, x) + b \right), 
	\ w = \sum_{j=1}^n \alpha_j \phi(x_j),
\end{align*}
where $\phi(x)$ is an only implicitly defined feature map (due to
Mercer's theorem, see also \cite{Vapnik:95}) corresponding to the
kernel, i.e., fulfilling $k(x, x') = \langle \phi(x), \phi(x') \rangle$,
and $x_1, \dots, x_n$ are the training points. Points $x_j$ with non-zero
coefficients $\alpha_j \not= 0$ are called support vectors; the summation
in the predictor can apparently be restricted to this subset. The weight
vector is obtained by minimizing
\begin{align}
	P(w, b) = \frac{\lambda}{2} \|w\|^2 + \frac{1}{n} \sum_{i=1}^n L \Big( y_i, \big\langle w, \phi(x_i) \big\rangle + b \Big).
	\label{eq:primal}
\end{align}
Here, $\lambda > 0$ is a user-defined regularization parameter and
$L(y, m) = \max\{0, 1 - y \cdot m\}$ denotes the hinge loss, which is a
prototypical large margin loss, aiming to separate the classes with a
functional margin $y \cdot m$ of at least one. By incorporating other
loss functions, SVMs can be generalized to other tasks like multi-class
classification \cite{dogan2016unified}.

\subsection{Stochastic Gradient Descent for Support Vector Machines}

Training an SVM with stochastic gradient descent (SGD) on the primal
optimization problem~\eqref{eq:primal} is similar to neural network
training. In most implementations including the Pegasos algorithm
\cite{Shalev-Shwartz:2007} input points are presented one by one, in 
random order. The objective function $P(w, b)$ is approximated by the
unbiased estimate
\begin{align*}
	P_i(w, b) = \frac{\lambda}{2} \|w\|^2 + L \Big( y_i, \big\langle w, \phi(x_i) \big\rangle + b \Big),
\end{align*}
where the index $i \in \{1, \dots, n\}$ follows a uniform distribution.
The stochastic gradient $\nabla P_i(w, b)$ is an unbiased estimate of
the ``batch'' gradient $\nabla P(w, b)$, however, it is faster to
compute by a factor of $n$ since it involves only a single training
point. Starting from $(w, b) = (0, 0)$, SGD updates the weights
according to
\begin{align*}
	(w, b) \leftarrow (w, b) - \eta_t \cdot \nabla P_{i_t}(w, b),
\end{align*}
where $t$ is the iteration counter. With a learning rate $\eta_t \in
\Theta(1/t)$ it is guaranteed to converge to the optimum of the convex
training problem \cite{Bottou:2010}. 

With the representation $w = \sum_{j=1}^n \alpha_j \phi(x_j)$, SGD
updates scale down the vector $\alpha = (\alpha_1, \dots, \alpha_n)$
uniformly by the factor $1 - \lambda \cdot \eta_t$. If the margin of
$(x_i, y_i)$ happens to be less than one, then the update also adds
$\eta_t \cdot y_i$ to $\alpha_i$. The most costly step is the
computation of $\langle w, \phi(x_i) \rangle$, which is linear in the
number of support vectors (SVs), and hence potentially linear in~$n$.
\subsection{Budget Stochastic Gradient Descent Algorithm (BSGD)}

BSGD breaks the unlimited growth in model size and update time for large
data streams by bounding the number of support vectors during training.
The upper bound $B \ll n$ is the budget size, a parameter of the method.
Per SGD step the algorithm can add at most one new support vector; this
happens exactly if $(x_i, y_i)$ violates the target margin of one and
$\alpha_i$ changes from zero to a non-zero value. After $B + 1$ such
steps, the budget constraint is violated and a dedicated budget
maintenance algorithm is triggered to reduce the number of support
vectors to at most~$B$. The goal of budget maintenance is to fulfill the
budget constraint with the smallest possible change of the model,
measured by $E = \|\Delta\|^2 = \|w' - w\|^2$, where $w$ is the weight
vector before and $w'$ is the weight vector after budget maintenance.
$\Delta = w' - w$ is referred to as the weight degradation.

Three budget maintenance strategies are discussed and tested by \cite{Wang:2012}: 
\begin{compactitem}
\item 
	removal of the SV with smallest coefficient $|\alpha_j|$,
\item
	projection of the removed SV onto the remaining SVs,
\item
	and merging of two SV to produce a new vector, which in general
	does not correspond to a training point.
\end{compactitem}
Removal was found to yields oscillations and poor results. Projection is
elegant but requires a computationally demanding $\Order(B^3)$ matrix
operation, while merging is rather fast and gives equally good results.
Merging was first proposed by \cite{Nguyen:2005} as a way to efficiently
reduce the complexity of an already trained SVM. Depending on the choice
of candidate pairs considered for merging, its time complexity is
$O(B^2)$ for all pairs and $O(B)$ if the heuristic of fixing the point
with smallest coefficient as a first partner is employed. Its good
performance at rather low cost made merging the budget maintenance
method of choice.

When merging two support vectors $x_i$ and $x_j$, we aim to approximate
$\alpha_i \cdot \phi(x_i) + \alpha_j \cdot \phi(x_j)$ with a single new
term $\alpha_z \cdot \phi(z)$ involving a single support vector $z$.
Since the kernel-induced feature map is usually not surjective, the
pre-image of $\alpha_i \phi(x_i) + \alpha_j \phi(x_j)$ under $\phi$ is
empty \cite{Schoelkopf:99,Burges:1996} and no exact match $z$ exists.
Therefore the weight degradation
$\Delta = \alpha_i \phi(x_i) + \alpha_j \phi(x_j) - \alpha_z \phi(z)$
is non-zero. For the Gaussian kernel
$k(x, x') = \exp(-\gamma \|x - x'\|^2)$, due to its symmetries, the
point $z$ minimizing $\|\Delta\|^2$ lies on the line connecting $x_i$
and $x_j$ and is hence of the form $z = h x_i + (1 - h) x_j$. For
$y_i = y_j$ we obtain a convex combination $0 < h < 1$, otherwise we
have $h < 0$ or $h > 1$. For each choice of $z$, the optimal value of
$\alpha_z$ can be obtained in closed form. This turns minimization of
$\|\Delta\|^2$ into a one-dimensional non-linear optimization problem,
which is solved by \cite{Wang:2012} with golden section search.

Budget maintenance in BSGD usually works as follows: $x_i$ is fixed to
the support vector with minimal coefficient $|\alpha_i|$. Then the best
merge partner $x_j$ is determined by testing $B$ pairs $(x_i, x_j)$,
$j \in \{1, \dots, B+1\} \setminus \{i\}$. The golden section search is
carried out for each of these in order to determine the resulting weight
degradation. Finally, the index $j$ with minimal weight degradation is
selected and the vectors are merged.

\section{Multi-Merge Budget Maintenance}

In this section we analyse the computational bottleneck of the BSGD
algorithm. Then we propose a modified \emph{multi-merge} variant that
addresses this bottleneck.

The runtime cost of a BSGD iteration depends crucially on two factors:
whether the point under consideration violates the target margin or not,
and if so, whether adding the point to the model violates the budget
constraint or not. Let $K$ denote the cost of a kernel computation
(which is assumed to be constant for simplicity). Then the computation
of the margin of the candidate point is an $\Order(B K)$ operation,
which is dominated by the computation of up to $B$ kernel values between
support vectors and the candidate point.

We consider the case that the target margin is violated and the point is
added to the model. According to \cite{steinwart2003sparseness} we can
expect this to happen for a fixed fraction of all training points. Since
a well-chosen budget size is significantly smaller than the number of
support vectors of the unconstrained solution (otherwise there is no
point in using a budget), we can assume that the initial phase of
filling up the budget is short, and the overall runtime is dominated by
iterations involving budget maintenance.

\begin{figure}
\begin{minipage}[t]{0.49\textwidth}
	\centering
	ADULT
	\includegraphics[width=\textwidth]{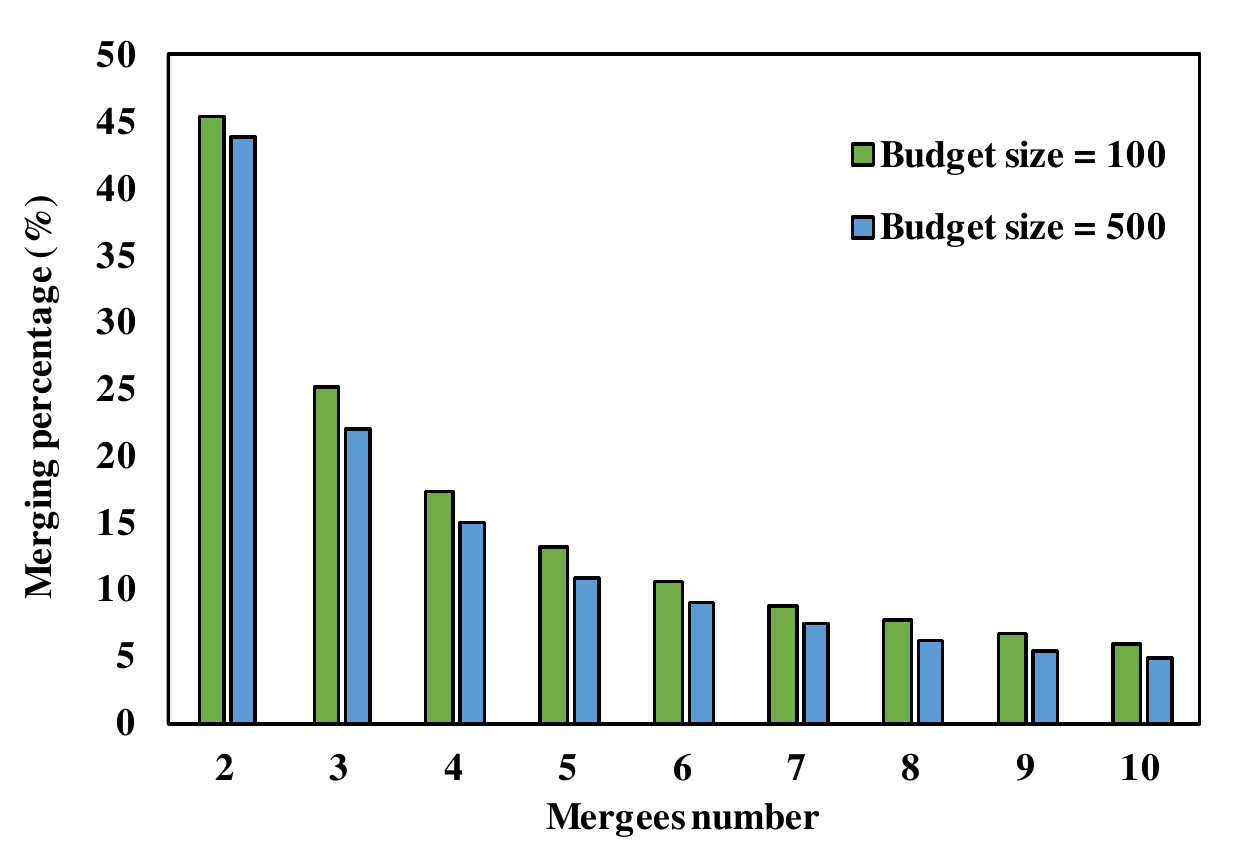}%
\end{minipage}%
\begin{minipage}[t]{0.49\textwidth}
	\centering
	IJCNN
	\includegraphics[width=\textwidth]{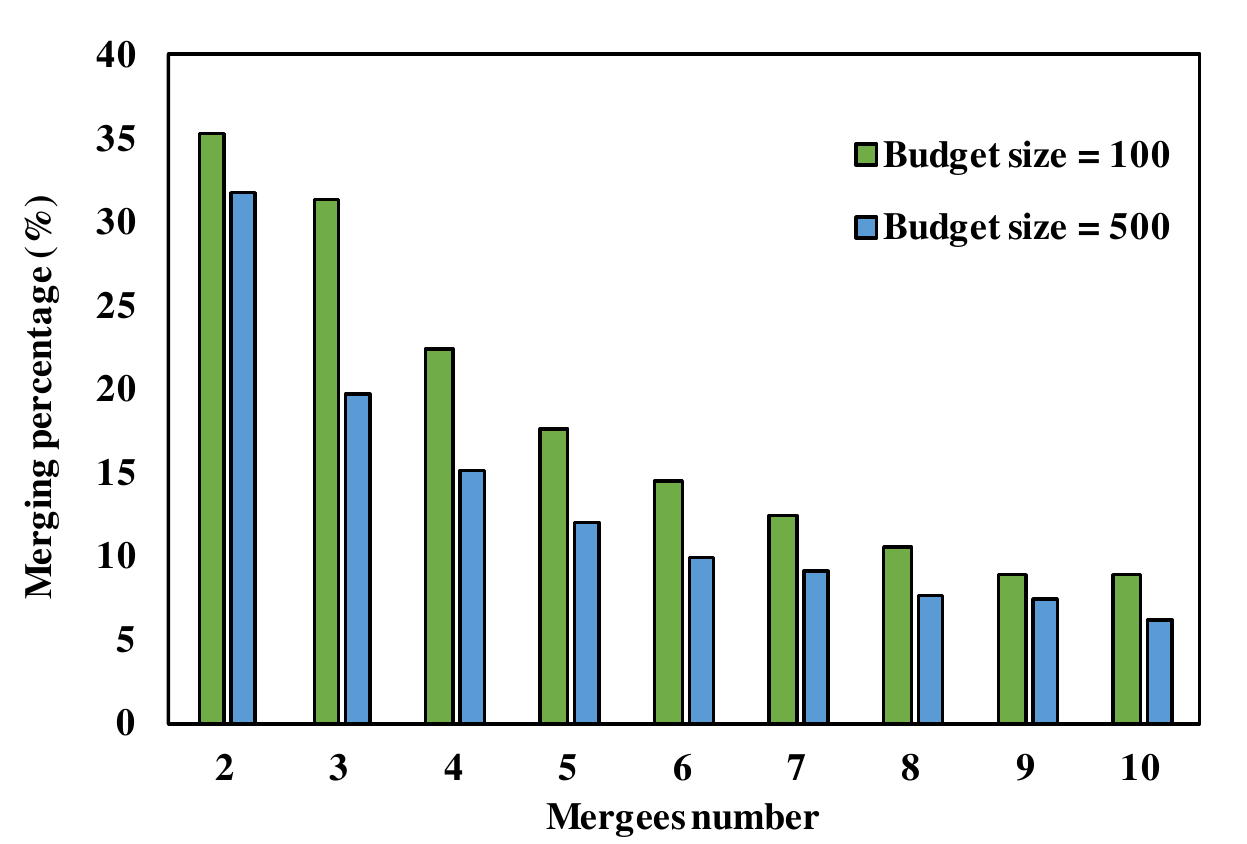}%
\end{minipage}%
\caption{\label{figure:figMergingAdult}
		Fraction of the overall training time spent on merging of
		support vectors, as a function of the number $M$ of mergees,
		for budget sizes $B=100$ and $B=500$ on the ADULT and  
		IJCNN data set.}
\end{figure}

During budget maintenance two support vectors are merged. Finding the
first merge candidate with minimal coefficient $|\alpha_i|$ is as cheap
as $\Theta(B)$. Finding the best merge partner requires solving $B$
optimization problems. Assuming a fixed number of $G$ iterations of the
golden section search takes $\Theta(B K G)$ operations. It turns
out empirically that the latter cost is significant; it often accounts
for a considerable fraction of the total runtime
(see figure~\ref{figure:figMergingAdult}, left-most case in both cases).
Therefore, in order to accelerate BSGD, we focus on speeding up the
budget maintenance algorithm.

When merging two support vectors, we aim to replace them with a single
point so that their weighted sum in feature space is well approximated
by the new feature vector. In principle, the same approach is applicable
to larger sets, i.e., merging $M$ into $m < M$ new points, and hence
reducing the current number of support vectors by $M - m$. The usual
choice $M = 2$ and $m = 1$ is by no means carved in stone. Due to the
need to reduce the number of support vectors, we stick to $m = 1$,
but we take the freedom to use a larger number $M$ of support vectors to
merge.

The main benefit of merging $M > 2$ points is that the costly budget
maintenance step is triggered only once for $M - 1$ training points that
violate the target margin. E.g., when merging $M = 5$ points at once,
we save 66\% of the budget maintenance calls compared to merging only
$M = 2$ points. This means that with large enough $M$, the fraction of
computation time that is spent on budget maintenance can be reduced,
and the lion's share of the time is spent on actual optimization steps.
However, for too large $M$ there may be a price to pay since merging
many points results in a large approximation error (weight degradation).
Also, the proceeding can only pay off if the computational complexity of
merging $M > 2$ points is not much larger than that of merging $M = 2$
points.

In order to obtain a speed-up, it is crucial not to increase the cost of
selecting the merge partners. In principle, when choosing $M$ points, we
select among all $\binom{B+1}{M}$ subsets. Already for $M = 2$ there are
$\Theta(B^2)$ pairs. Testing them all would increase the complexity by a
factor of $B$. Therefore the first point is picked with a heuristic,
reducing the remaining number of pairs to $B$. This heuristic extends to
larger sets, as follows.

Intuitively, good merge partners are close in input space, because only
then can we hope to approximate a sum of Gaussians with a single
Gaussian, and they have small coefficients, since a small contribution
to the overall weight vector guarantees a small worst case approximation
error. We conclude that the property of being good merge partners is
``approximately transitive'' in the following sense: if $y$ and $z$ are
both good merge partners for $x$, then their coefficients are small and
they are both close to $x$. Also, the triangle inequality
$\|y - z\| \leq \|y - x\| + \|z - x\|$ implies that $y$ and $z$ are not
too far from each other, and hence merging $y$ and $z$ is promising, and
so is merging all three points $x$, $y$ and $z$.

\begin{algorithm}[t]

	\caption{
		Decomposition of multiple merges into a sequence of binary
		merges (MM-BSGD)
		\label{MMBSGDI}
	}


	\SetKwInOut{Input}{Input}

	{select $SV_x$} 

	\Input{$M  \geq 3$}
	 {$(z, \alpha_z) = $ merge $((SV_1,  \alpha_1),(SV_2,  \alpha_2))$} 

	 \For{ $ (j \in \{3, \dots, M\} )$ }{
	 $(z, \alpha_z) = $ merge $((z, \alpha_z),(SV_j,  \alpha_j))$;
	}

\end{algorithm}

\begin{algorithm}[t]

	\caption{
		Weight degradation minimization through gradient descent (MM-GD)
		\label{MMGD}
	}


	\SetKwInOut{Input}{Input}

	\Input{$M \geq 3$ and $\epsilon > 0$}

	{select $SV_{1}, SV_{2}, ... , SV_{M}$}
    
	{initialize $z = \frac {\sum_{i=1}^{M} SV_i * \alpha_i}{ \sum_{i=1}^{M} \alpha_i} $ }

	{initialize $\alpha_{z} = { \sum_{i=1}^{M} \alpha_i}$ }

	{ $f_t  = \|\Delta_t \|^2 = min.  \| \sum_{i=1}^{M} \alpha_i \phi(x_i)  - \alpha_z \phi(z) \|^2\ $}

	\While { ( $\|\Delta_t \|$ $< \epsilon$ )}{
	 $z \leftarrow z -\eta_t\nabla_{z} (f_t)$\ ;
	 
	 optimize $\alpha_z$ ;
	  
	 calculate $\|\Delta_t \|$;
	}
\end{algorithm}

We turn the above heuristic into an algorithm as follows. The first
merge candidate is selected as before as the support vector with
smallest coefficient~$|\alpha_i|$. Then all other points are paired with
the first candidate, and the pairwise weight degradation are stored. The
best $M - 1$ points according to the criterion of minimal pairwise
weight degradation are picked as merge partners.%
\footnote{The selection of the $M - 1$ best points can be implemented in
  $\Theta(B)$ operations. However, in practice the $\Order(B \log(B))$
  operation of sorting by weight degradation does not incur a
  significant overhead, and it allows us to merge points in order of
  increasing weight degradation. For very large $B$ is may pay off to
  sort only the best $M - 1$ points.}
In this formulation, the only difference to the existing algorithm is
that we pick not only the single best merge partner, but $M - 1$ merge
partners according to this criterion.

Once $M > 2$ points are selected, merging them is non-trivial. With the
Gaussian kernel, the new support vector can be expected to lie in the
$M-1$ dimensional span of the $M$ points. Hence, golden section
search is not applicable. We propose two proceedings: 1.) minimization
of the overall weight degradation through gradient descent (MM-GD) and 2.) a
sequence of $M-1$ binary merge operations using golden section search (MM-BSGD).
The resulting multi-merge budget maintenance methods are given in
algorithm~\ref{MMGD} and algorithm~\ref{MMBSGDI}.

\cite{Wang:2012} provide a theoretical analysis of BSGD, in which the
weight degradation $\Delta$ enters as an input. Their theorem~1
(transcribed to our notation) read as follows:
\begin{theorem}[Theorem~1 by \cite{Wang:2012}]
Let $w^*$ denote the optimal weight vector (the minimizer of eq.~\eqref{eq:primal}),
let $w_t$ denote the sequence of weight vectors produced by (multi-merge)
BSGD with learning rate $\eta_t$, using the point with index $k_t$ in
iteration $t$. Let $E_t := \Delta_t / \eta_t$ denote the gradient error
and assume $\|E_t\| \leq 1$. Define the average gradient error as
$\bar E = \sum_{t=1}^N \|E_t\| / N$. Define $U = 2/\lambda$ if
$\lambda \leq 4$ and $U = 1/\sqrt{\lambda}$ otherwise.
Then the following inequality holds:
\begin{align*}
	\frac{1}{N} \sum_{t=1}^N P_{k_t}(w_t) - \frac{1}{N} \sum_{t=1}^N P_{k_t}(w^*) \leq \frac{(\lambda U + 2)^2 (\ln(N) + 1)}{2 \lambda N} + 2 U \bar E.
\end{align*}
\end{theorem}
The theorem applies irrespective of the type and source of $\Delta_t$,
and it hence applies in particular to multi-merge BSGD. We conclude that
our multi-merge strategy enjoys the exact same guarantees as the
original BSGD method.

\section{Experimental Evaluation}
\label{section:experiments}

We empirically compare the new budget maintenance algorithm to the
established baseline method of merging two support vectors. We would
like to answer the following questions:
\begin{compactenum}
\item
	How to merge $M > 2$ points---in sequence, or at once?
\item
	Which speed-up is achievable?
\item
	Which cost do we pay in terms of predictive accuracy, and which
	trade-offs are achievable?
\item
	How many support vectors should be merged?
\item
	How does performance depend on hyperparameter setting?
\end{compactenum}
\subsection{Merging Strategy}

We start with a simple experiment, aiming to clarify the first point.
We maintain the budget by merging $M = 3$ support vectors into one,
reducing the number of vectors by two in each budget maintenance step,
with algorithms \ref{MMBSGDI} $(3 \rightarrow 2 \rightarrow 1)$
and~\ref{MMGD} $(3 \rightarrow 1)$. The resulting training times and test
accuracies are found in table~\ref{tab:MergingWithGradientdescentANDgoldensection}
for a range of settings of the budget parameter~$B$. From the results we
can conclude that our gradient-based merging algorithm is a bit faster
than calling the existing method based on golden section search twice.
However, for realistic budgets yielding solid performance, differences
are minor, and in particular the test errors are nearly equal. We
conclude that the merging strategy does not have a major impact on the
results, and that simply calling the existing merging code $M-1$ times
when merging $M$ points is a valid approach. We pursue this strategy in
the following.
\begin{table}
  \centering
  \caption{
	Results of merging at $M = 3$ using gradient descent $(3\rightarrow 1)$ implemented in algorithm~\ref{MMGD}
	and in two cascaded steps $(3\rightarrow 2 \rightarrow 1)$ as implemented in algorithm \ref{MMBSGDI}
	for one epoch on the ADULT data set.
	\label{tab:MergingWithGradientdescentANDgoldensection}
  }
  \begin{tabular}{l l l l l l l}
     \toprule
     $B$        &      &   120  &   600  &   1200 &   1800 &   2500 \\
     \midrule
     Merging    & sec  & 10.562 & 27.222 & 53.362 & 79.896 & 109.922 \\
     $(3\rightarrow 2 \rightarrow 1)$ & (\%) & 76.32 & 82.97 & 83.36 & 84.04 & 83.98 \\[0.25em]
     Merging    & sec  & 6.042 & 25.982 & 50.641 & 75.560 & 96.686 \\
     $(3\rightarrow 1)$  & (\%) & 76.32 & 83.30 & 83.69 & 84.04 & 83.94 \\
 \bottomrule
  \end{tabular}
\end{table}

\subsection{Effect of Multiple Merges} \label{RC}

In order to answer the remaining three questions, we performed
experiments on the data sets PHISHING, WEB, ADULT, IJCNN, and
SKIN/NON-SKIN, covering a spectrum of data set sizes. For simplicity we
have restricted our selection to binary classification problems.
The SVM's hyperparameters, i.e., the complexity control parameter~$C$
and the bandwidth parameter~$\gamma$ of the Gaussian kernel (which was
used in all experiments), were tuned with grid search and
cross-validation. Data set statistics, hyperparameter settings, and
properties of the ``exact'' LIBSVM solution are summarized in
table~\ref{tab:tableproperties}.
\begin{table}[h!]
  \centering
  \caption{
    Data sets used in this study, SVM's hyperparameters $C$ and $\gamma$ tuned with grid search and cross-validation, and properties of the ``exact'' solution found by LIBSVM.
    \label{tab:tableproperties}
  }
   \begin{tabular}{p{2cm} l l l l l }
    \toprule
    
    data set  & size    & \# features & $C$ & $\gamma$  & test~accuracy \\
      \midrule
    PHISHING  &   8,315 &           68 &   8 &     8     & 97.55\% \\
    WEB       &  17,188 &         300 &    8 &     0.03  & 98.80\% \\
    ADULT     &  32,561 &         123 &   32 &     0.008 & 84.82\% \\
    IJCNN     &  49,990 &          22 &   32 &     2     & 98.77\% \\
    SKIN      & 164,788 &           3 &    8 &     0.03  & 98.96\% \\
 \bottomrule
  \end{tabular}
\end{table}

\begin{figure*}%
	\centering
	\includegraphics[width=\textwidth]{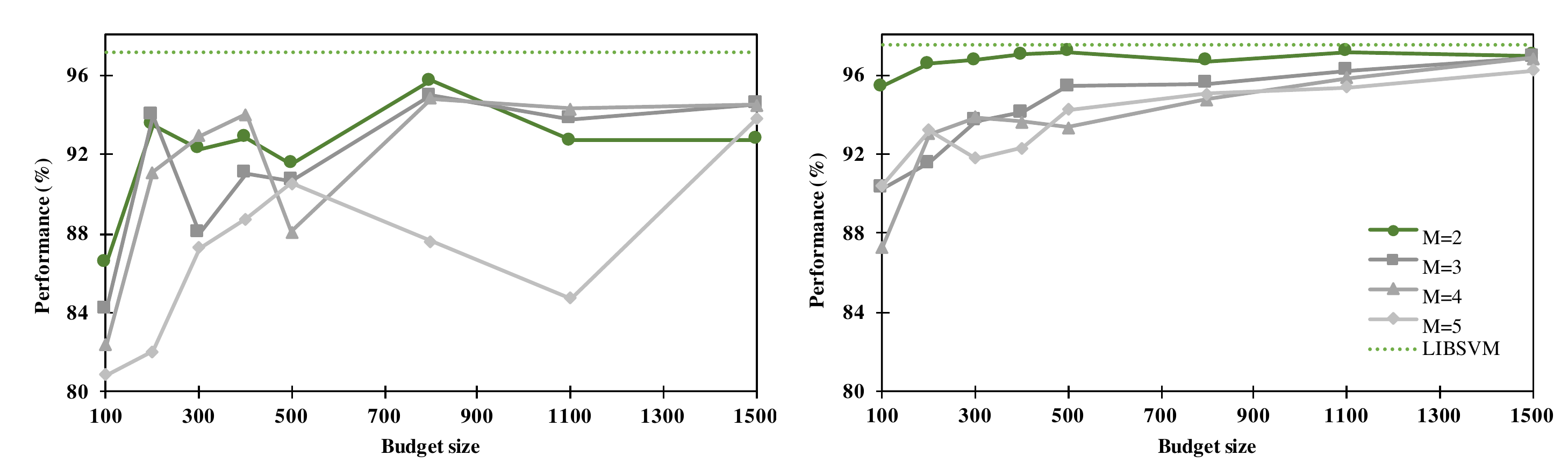}\\
	\includegraphics[width=\textwidth]{web}\\
	\includegraphics[width=\textwidth]{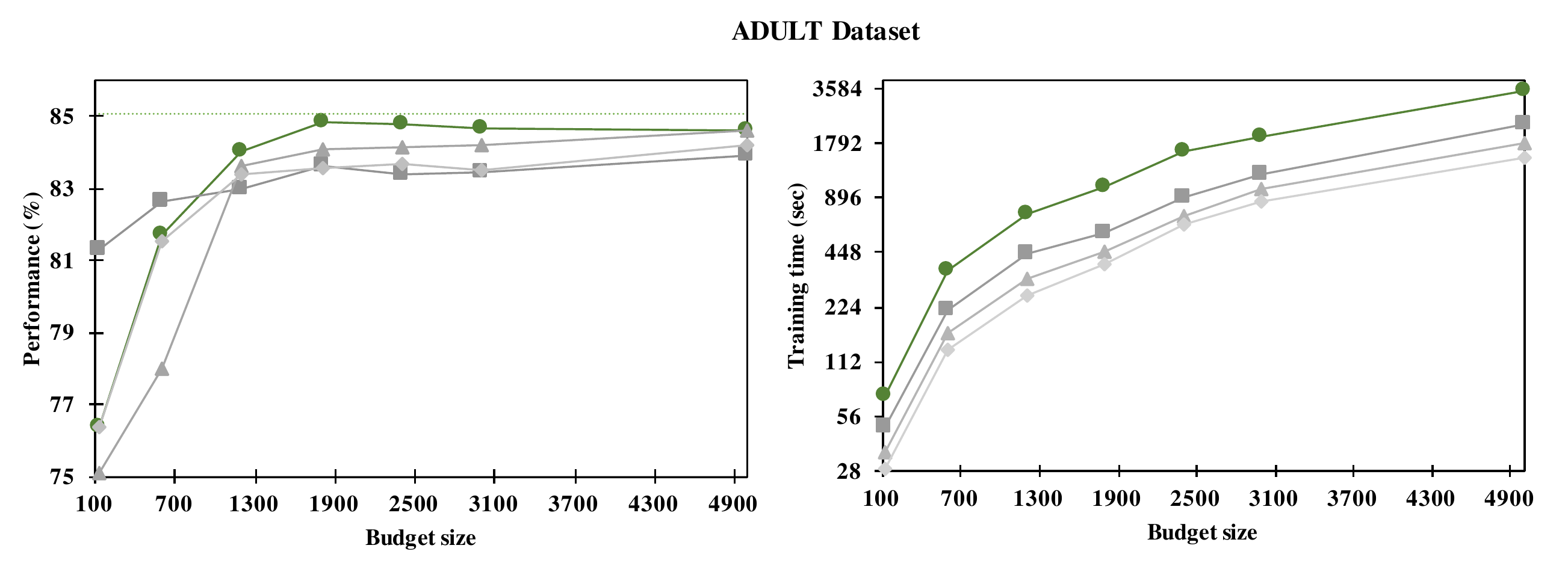}\\
	\includegraphics[width=0.75\textwidth]{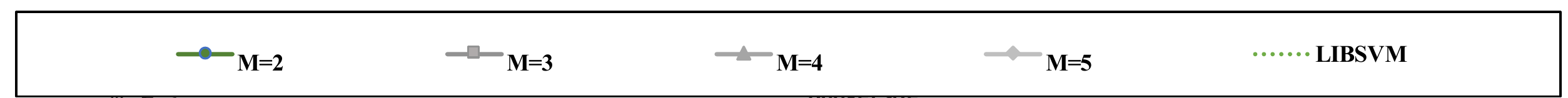}
	\caption{
	 Test accuracy (left) and training time (right) of the multi-merge
	 algorithm for the data sets PHISHING, WEB, and ADULT
	 (top to bottom), for different budget sizes $B$ and mergees $M$.
	 The test accuracy of the ``full'' SVM model obtained with LIBSVM is
	 indicated as a horizontal dotted line for reference. Notice the
	 logarithmic scale of the (vertical) time axis.
	 \label{figure:TimeAndAccuracy:1}
	}
\end{figure*}

\begin{figure*}%
	\centering
	\includegraphics[width=\textwidth]{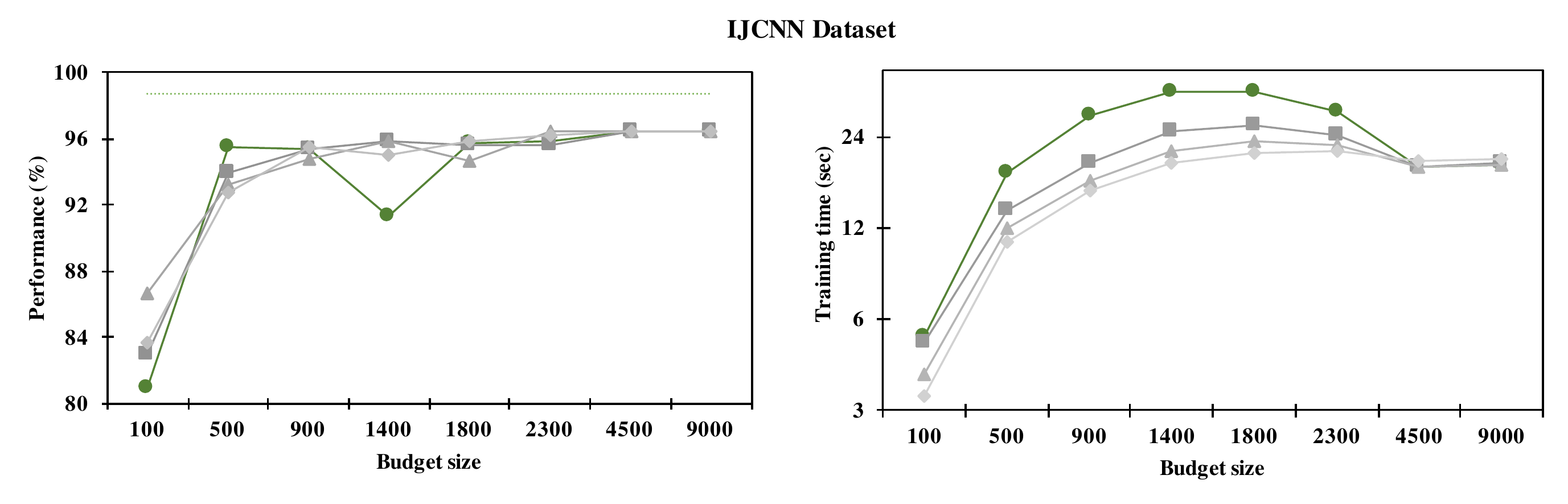}\\
	\includegraphics[width=\textwidth]{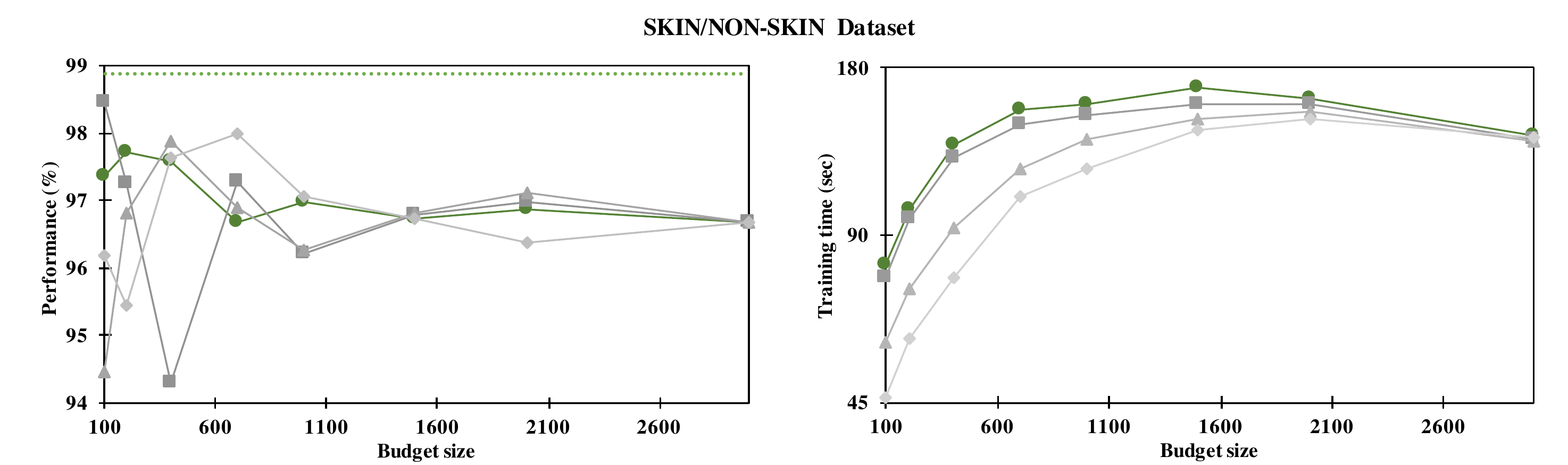}\\
	\includegraphics[width=0.75\textwidth]{legend}
	\caption{
	 Test accuracy (left) and training time (right) of the multi-merge
	 algorithm for the data sets IJCNN and SKIN/NON-SKIN (top to bottom),
	 for different budget sizes $B$ and mergees $M$.
	 The test accuracy of the ``full'' SVM model obtained with LIBSVM is
	 indicated as a horizontal dotted line for reference. Notice the
	 logarithmic scale of the (vertical) time axis.
	 \label{figure:TimeAndAccuracy:2}
	}
\end{figure*}

We are mainly interested in the effect of the number $M$ of merges.
This parameter was therefore systematically varied in the range
$M \in \{2, \dots 11\}$, corresponding to 1 to 10 binary merge
operations per budget maintenance. The minimal setting $M=2$ corresponds
to the original algorithm. Since the results may depend on the budget
size, the budget parameter~$B$ was varied as a fraction of roughly
$\{1\%, 5\%, 10\%, 15\%, 25\%,  50\%\}$ of the number of support
vectors of the full SVM model, as obtained with LIBSVM \cite{Chang:2011}. 
We have implemented our algorithm into the existing BSGD reference
implementation by Wang et al.~\cite{Wang:2012}. The results for
$M \in \{2, 3, 4, 5\}$ are shown in figures \ref{figure:TimeAndAccuracy:1}
and \ref{figure:TimeAndAccuracy:2}.

The results show a systematic reduction of the training time, which
depends solely on $M$, unless the budget is so large that budget
maintenance does not play a major role---however, in that case it may be
advisable to train a full SVM model with a dual solver instead. The
corresponding test accuracies give a less systematic picture. As
expected, the performance is roughly monotonically increasing with~$B$.
Results for small $B$ are somewhat unstable due to the stochastic nature
of the training algorithm.

We do not see a similar monotonic relation of the results w.r.t.~$M$.
On the ADULT, IJCNN and SKIN/NON-SKIN data sets, larger values of $M$
seem to have a deteriorative effect, while the contrary is the case for
the WEB problem. However, in most cases the differences
are rather unsystematic, so at least parts of the differences can be
attributed to the randomized nature of the BSGD algorithm. 
Overall, we do not find a systematic deteriorative effect due to
applying multiple merges, while the training time is reduced
systematically.

\begin{figure*}%
	\centering
	\includegraphics[width=\textwidth]{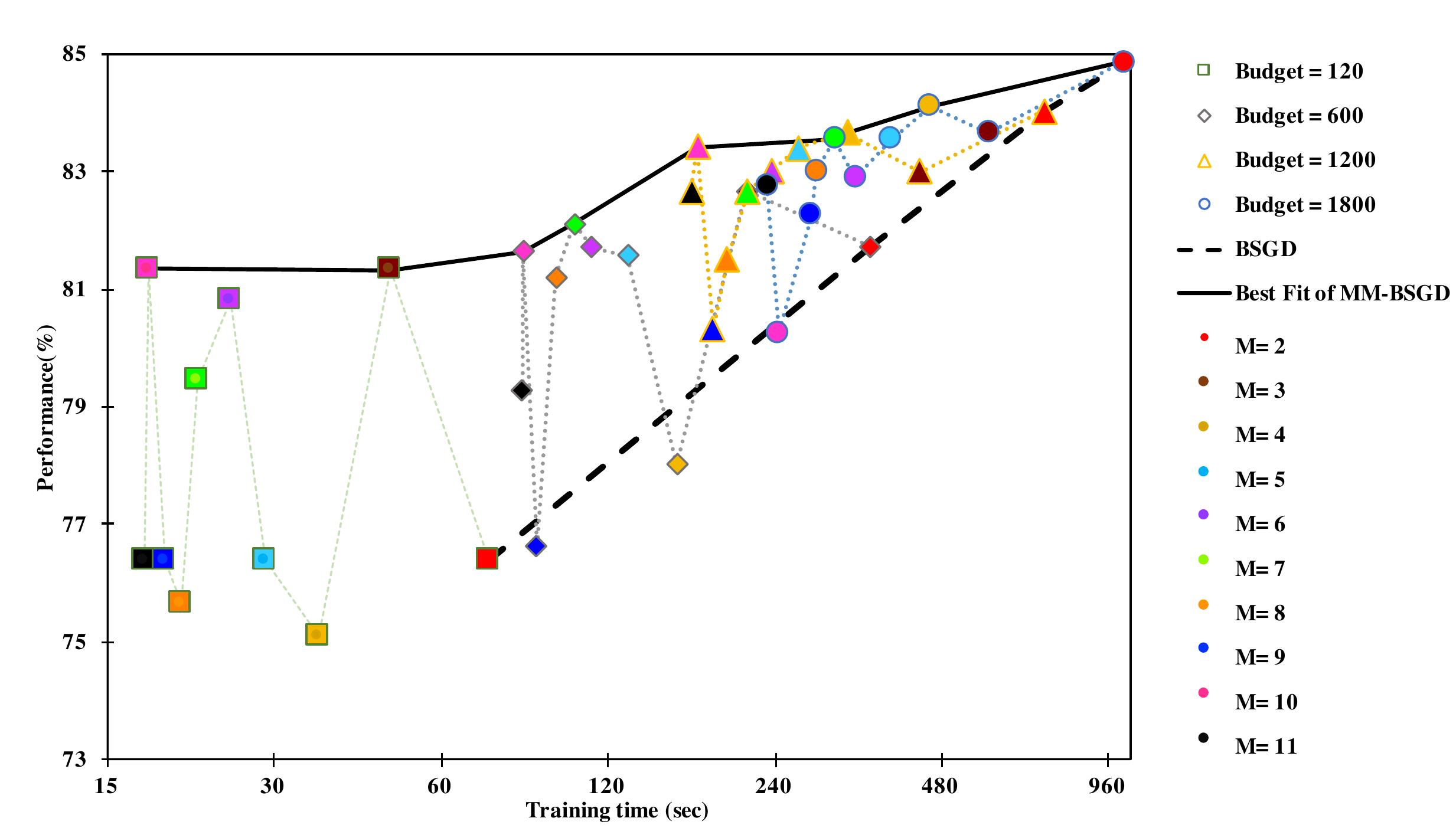}
	\caption{
	\label{figure:Pareto}
		Trade-off between accuracy and training time for the ADULT data
		set. The value of $M$ is color coded, and in addition, points
		corresponding to the same budget are connected with thin dotted
		lines. Different budgets are encoded with different symbols. The
		Pareto front of best trade-offs is depicted as a solid black
		line, connecting the non-dominated solutions marking optimal
		trade-offs between accuracy and training time.
	}
\end{figure*}

The same data is visualized differently in Figure~\ref{figure:Pareto},
which focuses on the achievable trade-offs between accuracy and training
time. This figure is based on the ADULT data, where our method does not
seem to perform very well according to Figures
\ref{figure:TimeAndAccuracy:1} and \ref{figure:TimeAndAccuracy:2}.
The figure displays accuracy over training time for a variety of
budgets~$B$ and mergees~$M$.

It is remarkable that all training runs with $M=2$ (black dashed line)
are found opposite to the Pareto front, i.e., they correspond to the
worst performance. The only exception appears for the largest budget
size, where training takes longest (which is exactly what we want to
avoid by using a budget in the first place), however, resulting in the
most accurate predictor in the field. The decisive observation is that
when starting with the baseline method it usually pays off to merge more
points in order to speed up the algorithm. A possible drop in
performance can be compensated by increasing the budget, resulting in a
simultaneous improvement of speed and accuracy over the baseline.

On the other hand, it is hard to conclude from the results which
strategy to use. Setting of $M=6$, $M=7$ and $M=9$ seem to give good
results, however, this effect should not be overrated, since we see some
not so good results for $M=5$, $M=8$ and also for $M=10$. 

In an attempt to answer how many support vectors should be merged, we
therefore recommend to use $M \in \{3, 4, 5\}$ in practice, since
smaller values can be expected to deliver stable results, while they
already realize a significant reduction of the training time. This
saving should then be re-invested into a larger budget, which in turn
results in improved accuracy.

\subsection{Parameter Study}

It is well known that the effort for solving the SVM training problem to
a fixed precision depends on the hyperparameters. For example, a very
broad Gaussian kernel (small~$\gamma$) yields an ill-conditioned kernel
matrix, and a large value of $C$ implies a large feasible region for
the dual problem, resulting in a solution with many free variables, the
fine tuning of which is costly with dual decomposition techniques. The
combination of both effects can result in very long training times, even
for moderately sized data sets.
For budget methods a major concern is the approximation error due to the
budget constraint. This error can be expected to be large if the kernel
is too peaked (large~$\gamma$).

Against this background it is unsurprising that also the efficiency of
the multi-merge method varies with hyperparameter settings. When merging
two points with a distance larger than about $\sqrt{1/\gamma}$, weight
degradation is unavoidable, and merging may simply result in a removal
of the point with smaller weight, which is known to lead to oscillatory
behavior and poor models. Such events naturally become more frequent for
large values of $\gamma$. In this case, merging even more points could
possibly be detrimental.

\begin{figure*}[ht]%
\centering
\footnotesize
\begin{minipage}[t]{0.33\textwidth}
	\centering
	$C=2$, $\gamma=2$, acc: $96.3490\%$,\\ \#SV: $1,367$
	\includegraphics[width=\textwidth]{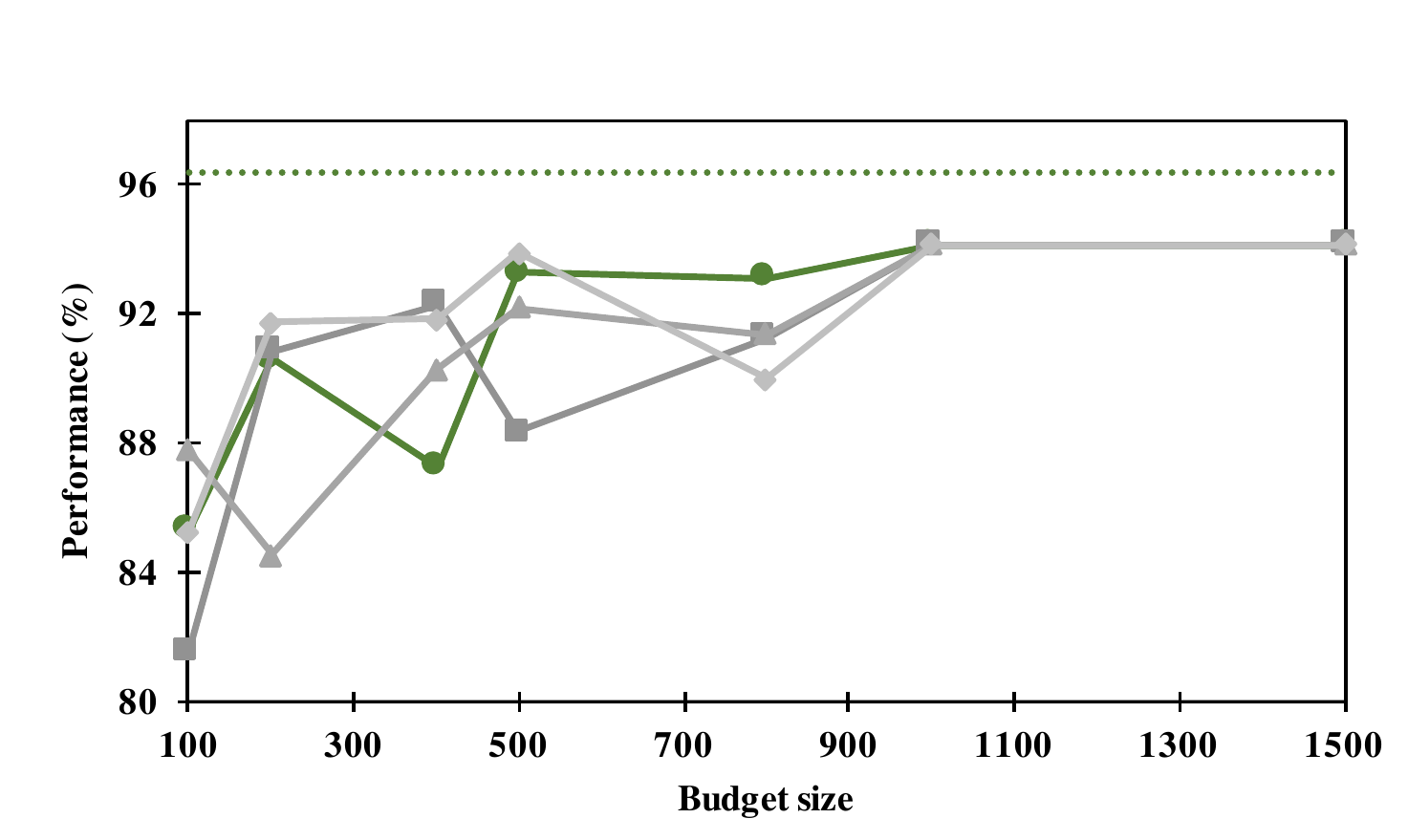}%
\end{minipage}%
\begin{minipage}[t]{0.33\textwidth}
	\centering
	$C=2$, $\gamma=8$, acc: $97.5904\%$, \\ \#SV: $2,852$
	\includegraphics[width=\textwidth]{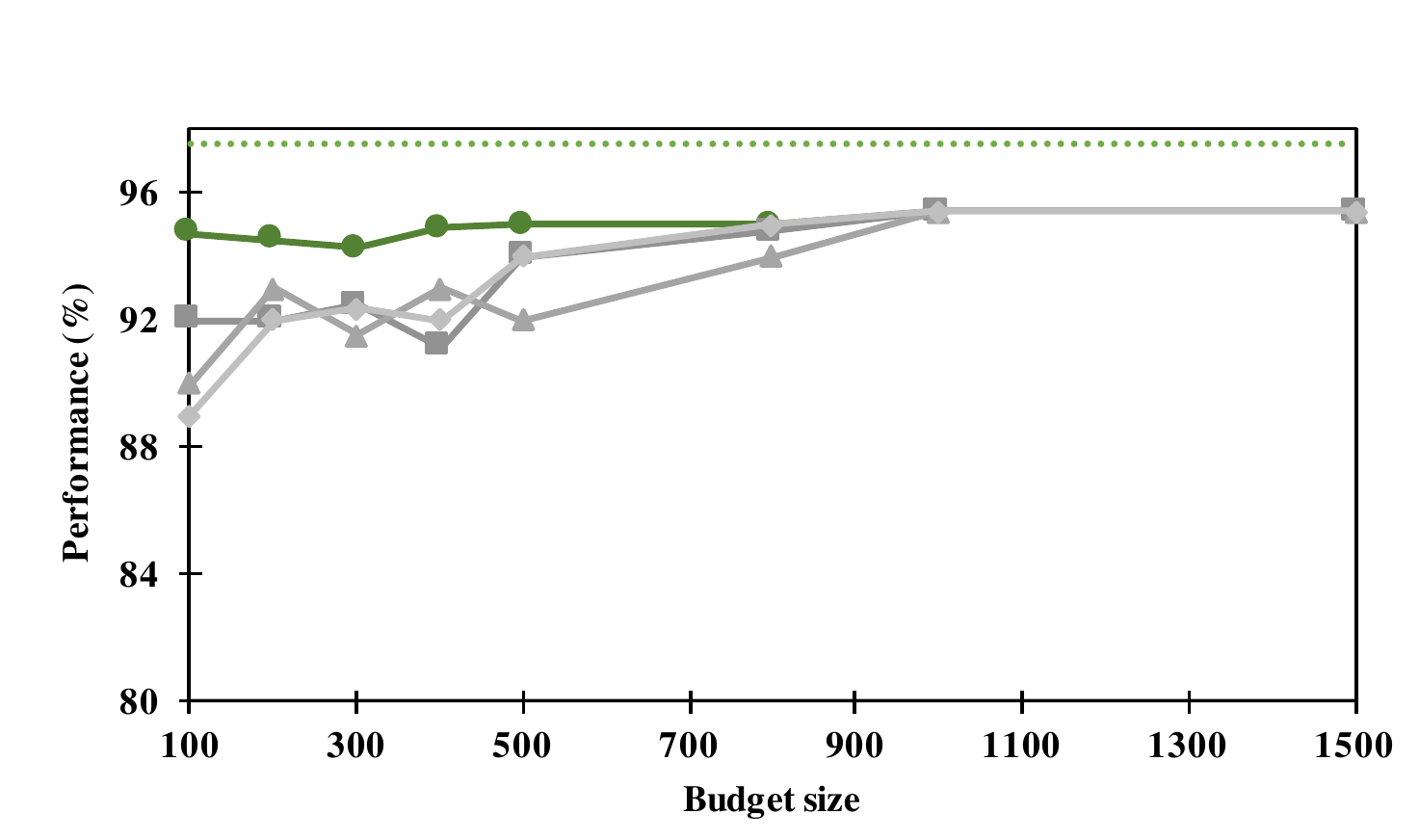}%
\end{minipage}%
\begin{minipage}[t]{0.33\textwidth}
	\centering
	$C=2$, $\gamma=32$, acc: $97.2983\%$, \#SV: $5,073$\\
	\includegraphics[width=\textwidth]{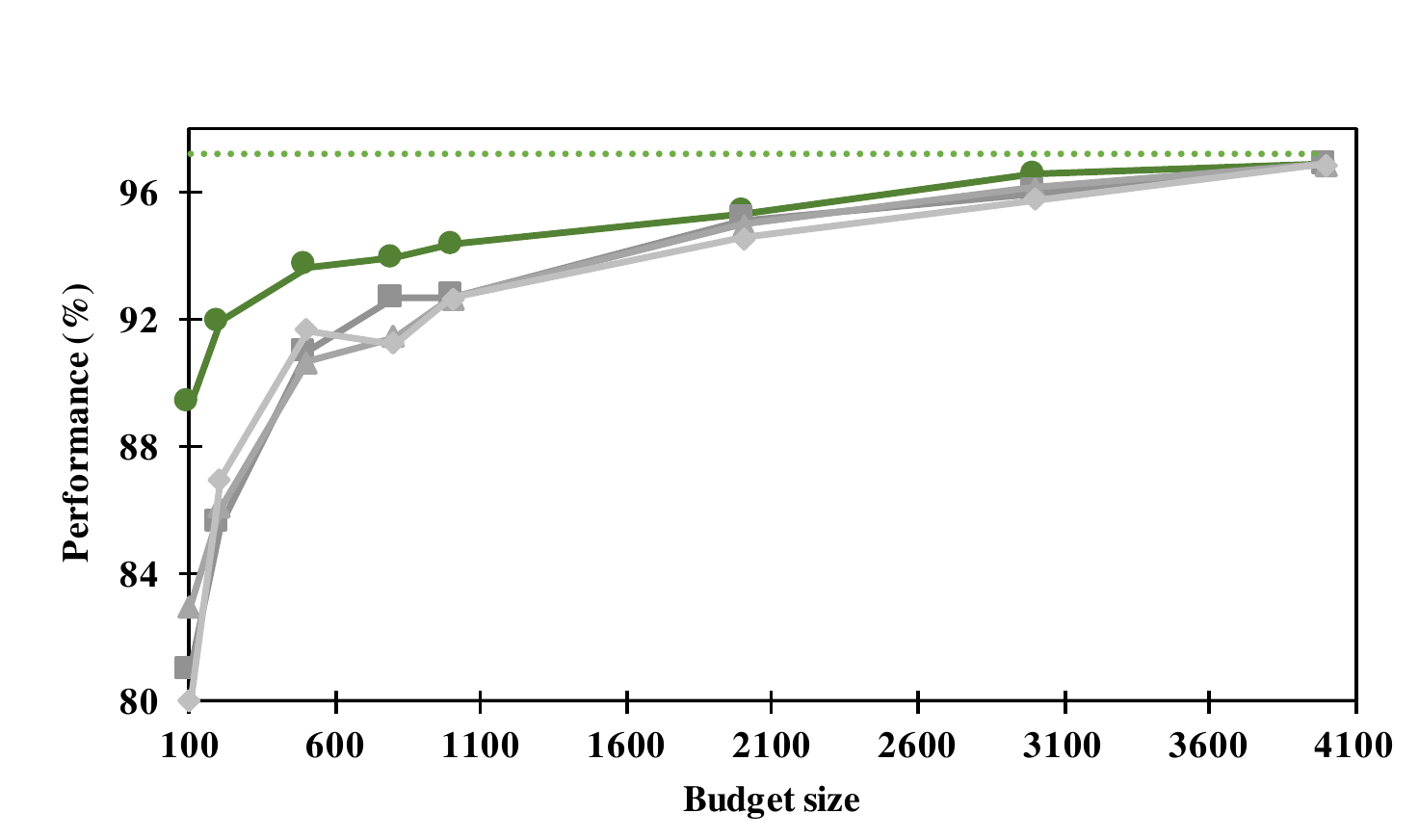}%
\end{minipage}%
\\[4ex]
\begin{minipage}[t]{0.33\textwidth}
	\centering
	$C=8$, $\gamma=2$, acc: $96.9697\%$,\\  \#SV: $1,191$
	\includegraphics[width=\textwidth]{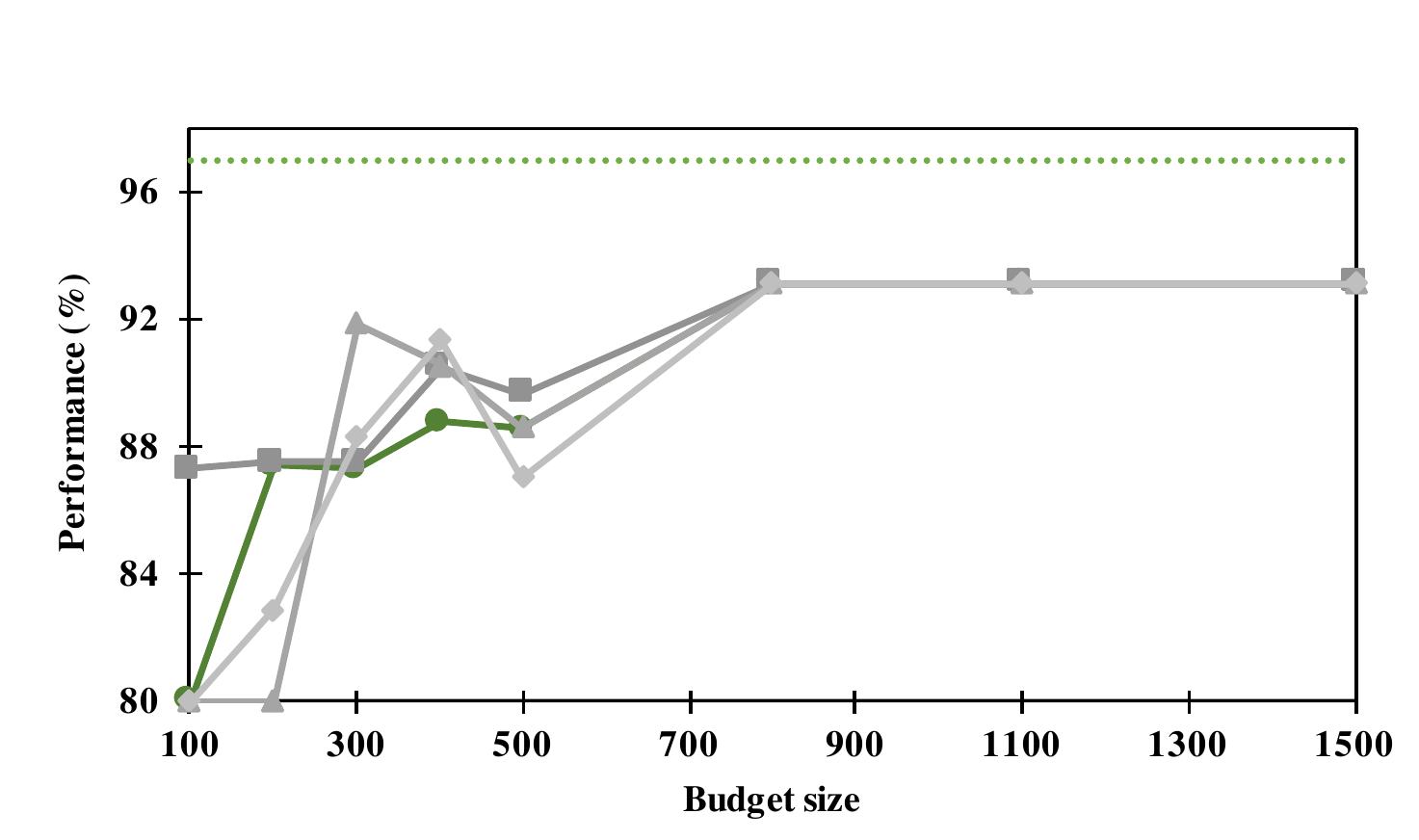}%
\end{minipage}%
\begin{minipage}[t]{0.33\textwidth}
	\centering
	$C=8$, $\gamma=8$, acc: $97.5173\%$,\\  \#SV: $2,751$
	\includegraphics[width=\textwidth]{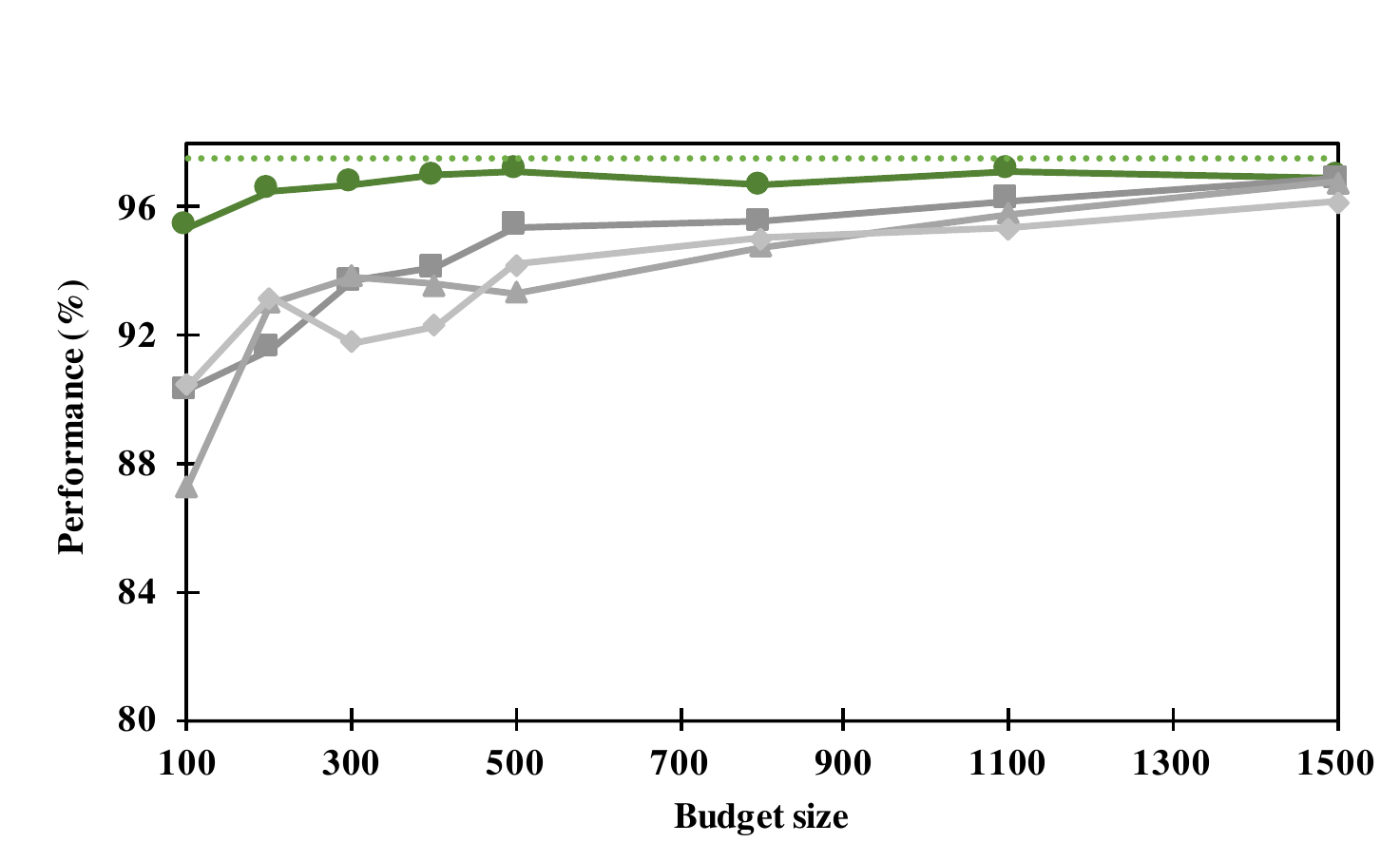}%
\end{minipage}%
\begin{minipage}[t]{0.33\textwidth}
	\centering
	$C=8$, $\gamma=32$, acc: $97.2983\%$,\\ \#SV: $5,073$
	\includegraphics[width=\textwidth]{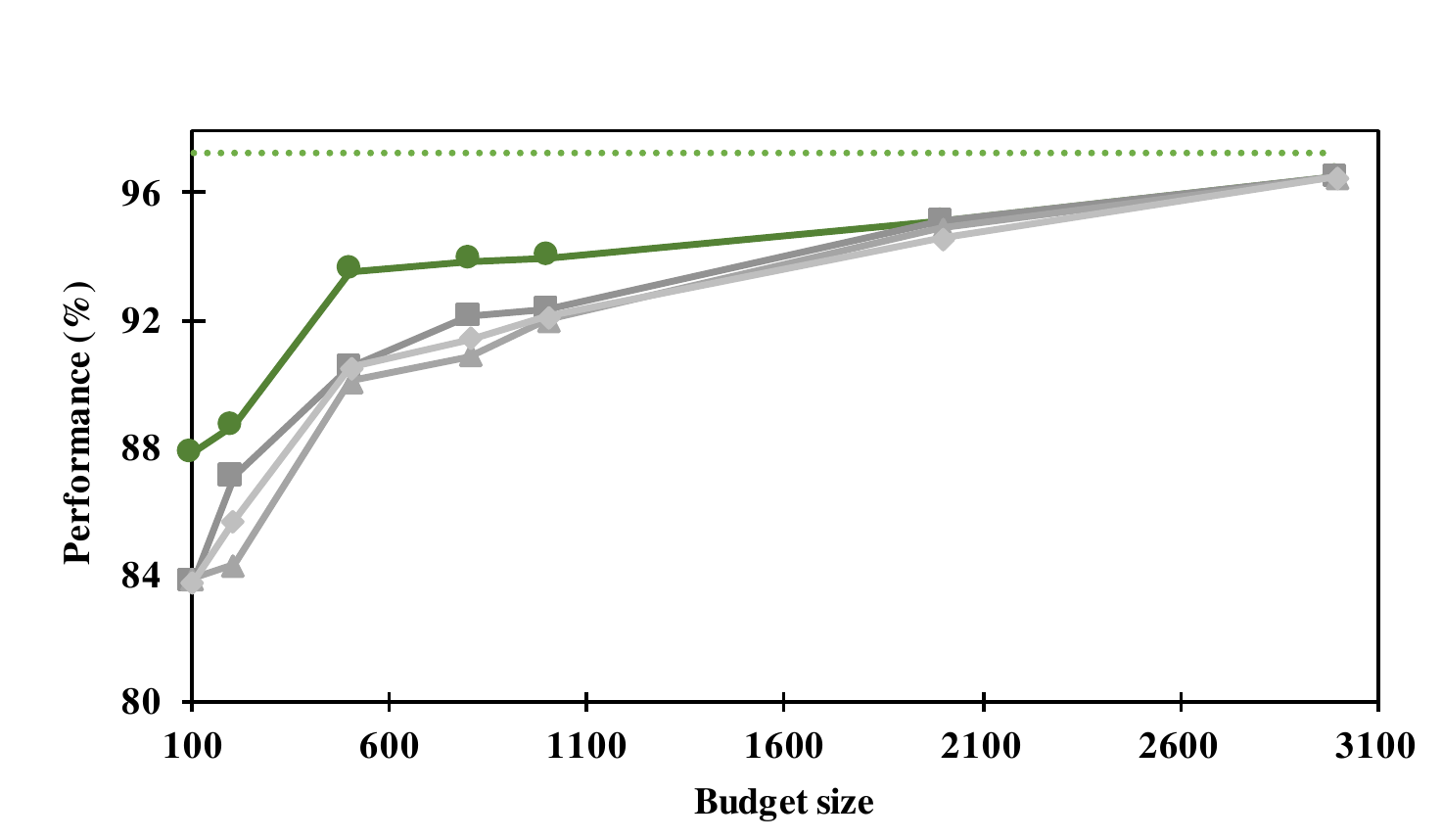}%
\end{minipage}%
\\[4ex]
\begin{minipage}[t]{0.33\textwidth}
	\centering
	$C=32$, $\gamma=2$, acc: $97.1522\%$,\\ \#SV: $1,106$
	\includegraphics[width=\textwidth]{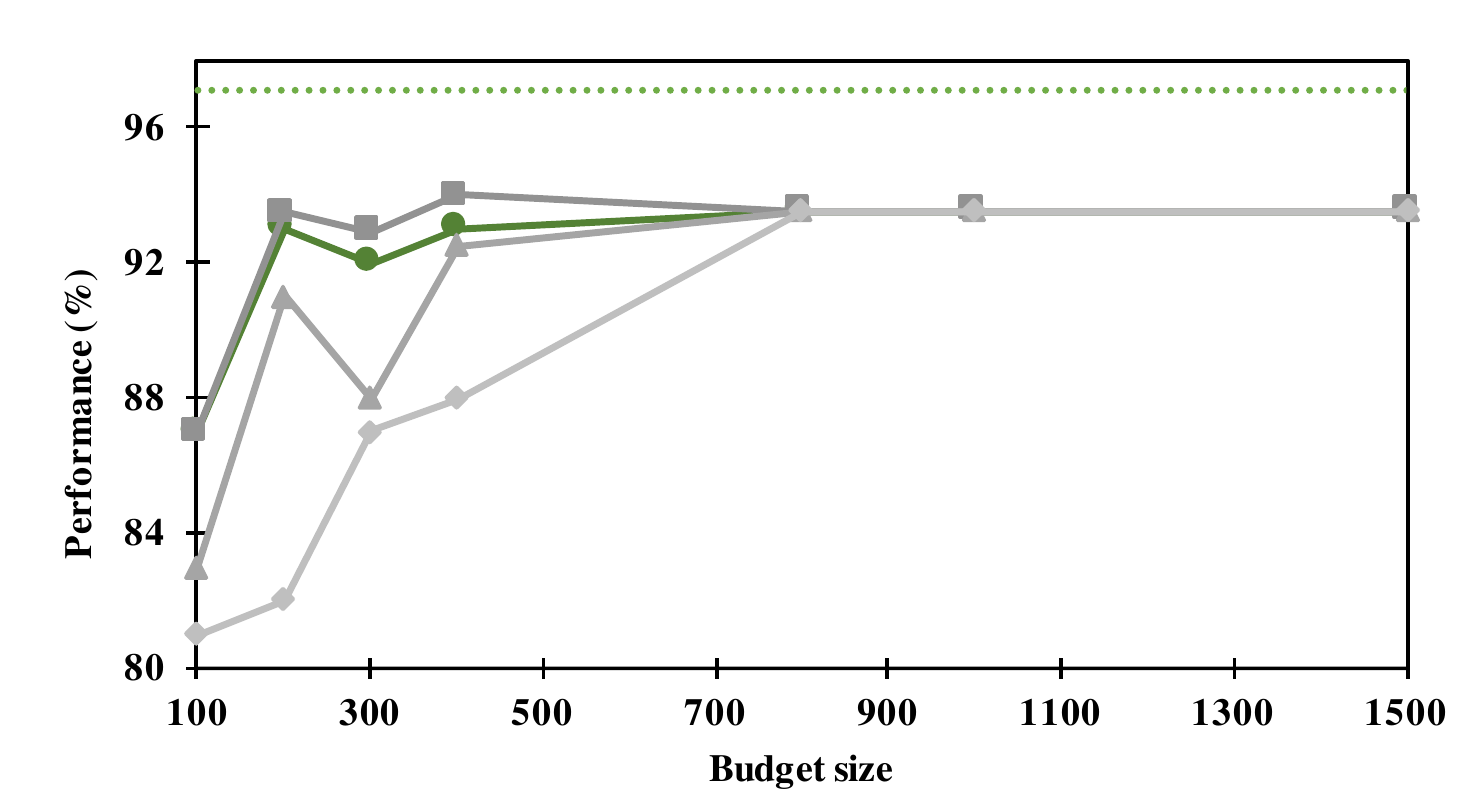}%
\end{minipage}%
\begin{minipage}[t]{0.33\textwidth}
	\centering
	$C=32$, $\gamma=8$, acc: $97.4804\%$, \#SV: $2,756$\\
	\includegraphics[width=\textwidth]{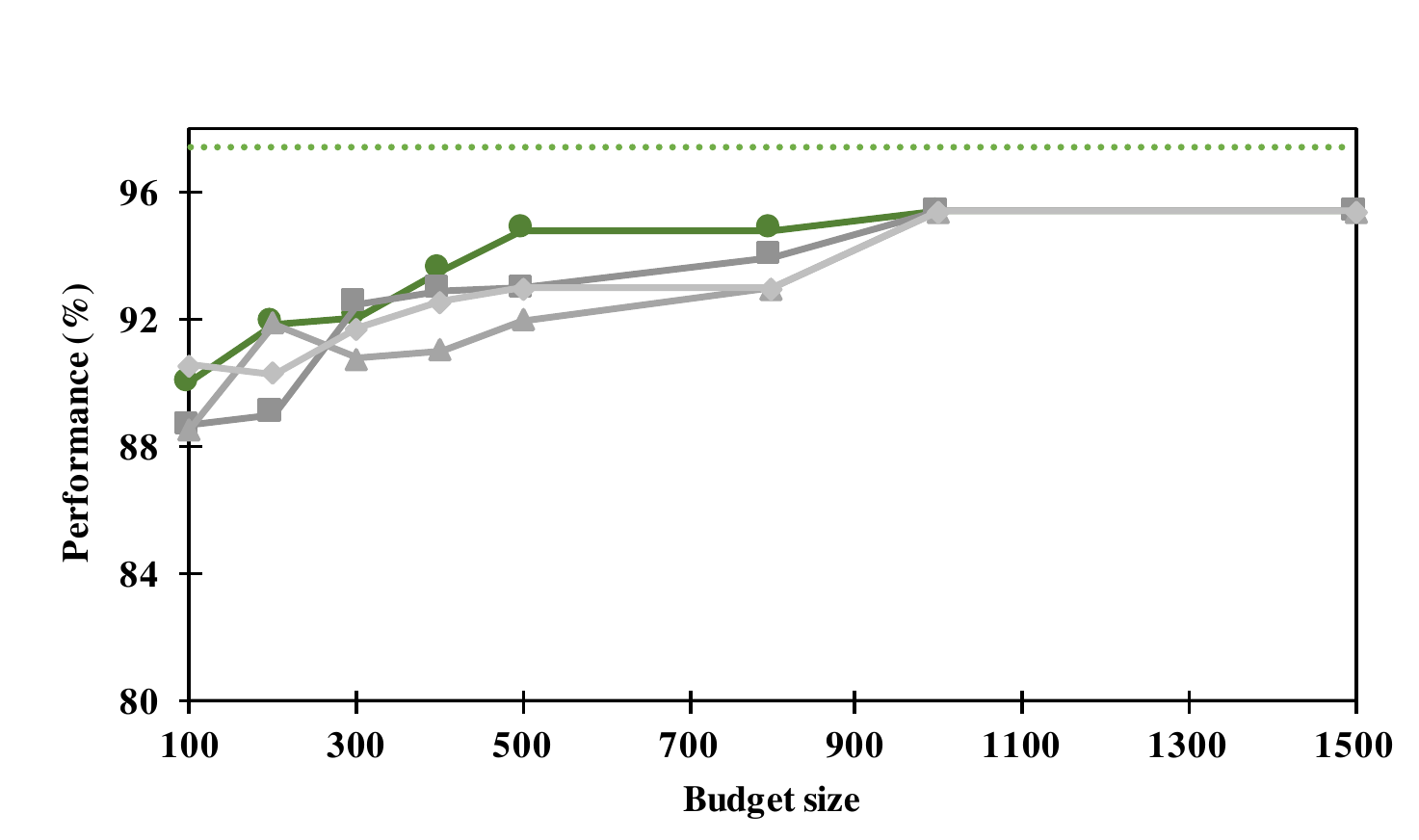}%
\end{minipage}%
\begin{minipage}[t]{0.33\textwidth}
	\centering
	$C=32$, $\gamma=32$, acc: $97.2983\%$, \#SV: $5,073$\\
	\includegraphics[width=\textwidth]{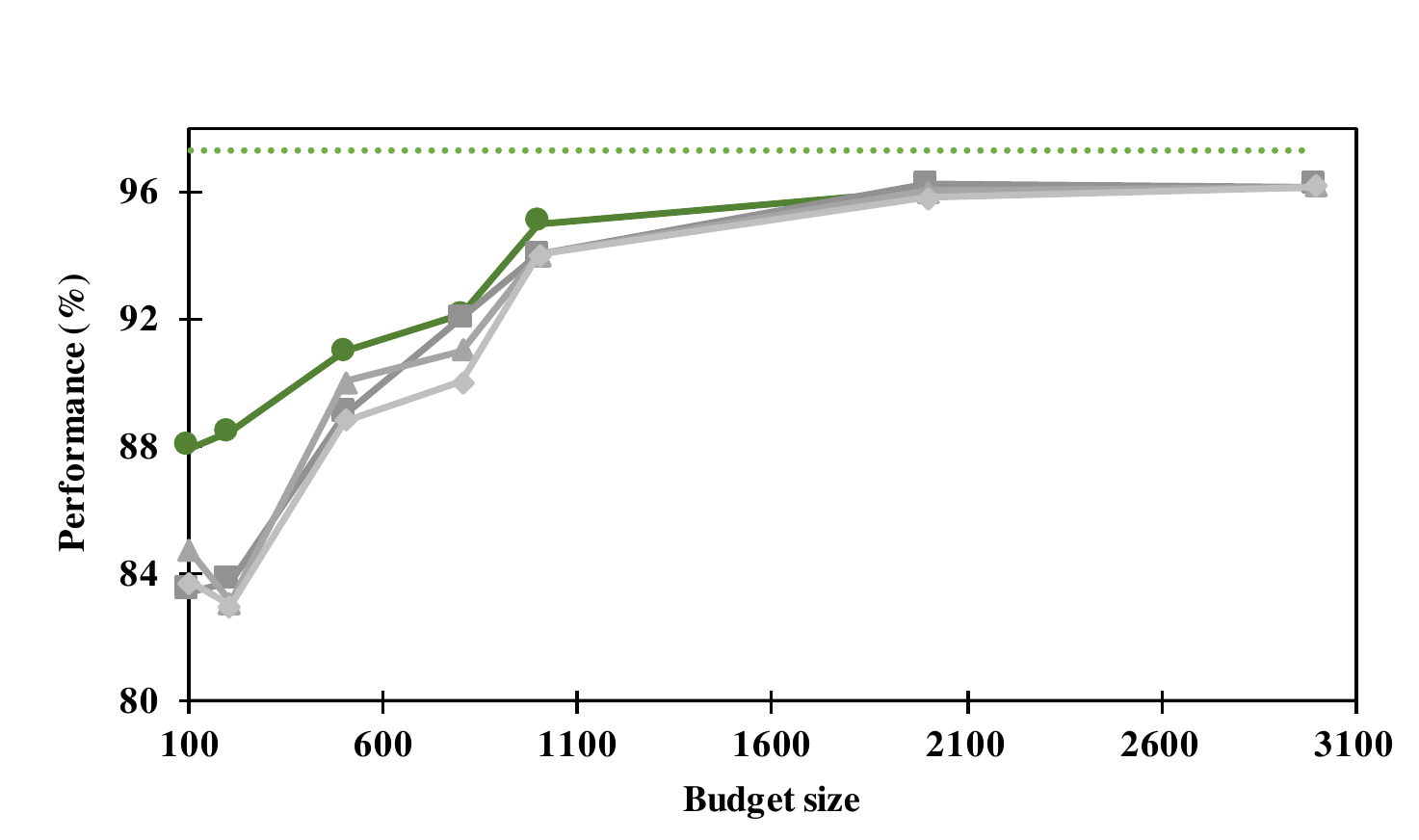}%
\end{minipage}%
\\
\includegraphics[width=0.75\textwidth]{legend}
\caption{
	Accuracy of LIBSVM (dashed), plain BSGD (green), and the multi-merge
	strategy with $M \in \{3, 4, 5\}$ for systematically varied
	hyperparameter settings and budget size on the PHISHING data set.
	\label{figure:phishingFS}
}
\end{figure*}

In order to clarify this question, we have varied the SVM
hyperparameters $C$ and $\gamma$ systematically.
Figure~\ref{figure:phishingFS} shows the results for nine parameter
configurations for the PHISHING data set, where the best known
configuration from table~\ref{tab:tableproperties} is in the center.
Note that the budgets differ for different values of $\gamma$, tracking
the numbers of support vectors of the LIBSVM model. In all cases, the
relevant range of budgets for which an actual time saving is achieved
covers roughly the leftmost $30\%$ of the plot.

It is immediate that the effect of changing $\gamma$ is much stronger
than that of changing $C$. The most prominent effect is that random
fluctuations of the performance are most pronounced for small $\gamma$.
This is expected, since in this regime the Gaussian kernel matrix is
badly conditioned, resulting in the hardest optimization problems.
Running SGD here only for a single epoch gives unreliable results,
irrespective of the presence or absence of a budget, and independent of
the budget maintenance strategy. This should get even worse for larger
values of $C$, corresponding to less regularization, but the effect is
completely masked by differences in performance due to different
hyperparameters, as well as noise, since we are working with a
randomized algorithm. Furthermore, as expected results stabilize and
improve as the budget increases.

Aside from the rather clear trends that reflect properties of the SVM
training problem and not of the algorithm, it is hard to find any
systematic effect of hyperparameter settings on the performance of our
multi-merge algorithm. We suspect that the expected effect is largely
masked by the above described more prominent factors. To answer our last
question, we conclude that our method works equally well across a range
of hyperparameter settings.

\section{Conclusions}

We have proposed a new multi-merge (MM) strategy for maintaining the
budget constraint during support vector machine training in the BSGD
framework. It offers significant computational savings by executing the
search for merge partners less frequently. Our method allows the BSGD
algorithm to spend the lion's share of its computation time on the
actual optimization steps, and to reduce the overhead due to budget
maintenance to a small fraction.
Experimental results show that merging more than two points at a time
yields significant speed-ups without degrading prediction performance as
long as a moderate number of points is merged. The best re-investment of
the reduced training time seems to be an increase of the budget size,
which in turn yields more accurate predictors. The results indicate that
with this strategy, highly accurate models can be trained in shorter
time.

\section*{Acknowledgments}

We acknowledge support by the Deutsche Forschungsgemeinschaft (DFG)
through grant GL~839/3-1.

\bibliographystyle{plain}

\end{document}